# Emotion-Aware Embedding Fusion in LLMs (Flan-T5, LLAMA 2, DeepSeek-R1, and ChatGPT 4) for Intelligent Response Generation

Abdur Rasool[1,*], Muhammad Irfan Shahzad[2], Hafsa Aslam[3], Vincent Chan[1], Muhammad Ali Arshad[4]

[1]Department of Information and Computer Sciences, University of Hawaii at Manoa, Honolulu, HI 96822, USA
[2]SelTeq, University Avenue, Palo Alto, CA 94301, USA (Email: ishah@selteq.net)
[3]School of Computer Science and Technology, Donghua University, Shanghai, China (Email: 324091@mail.dhu.edu.cn)
[4]Department of Computer Science and Technology, Nanjing University of Aeronautics and Astronautics, Nanjing 210016, China (Email: m.aliarshad108@nuaa.edu.cn)

Corresponding Author Email: abdur@hawaii.edu

## Abstract

Empathetic and coherent responses are critical in automated chatbot-facilitated psychotherapy. This study addresses the challenge of enhancing the emotional and contextual understanding of large language models (LLMs) in psychiatric applications. We introduce Emotion-Aware Embedding Fusion, a novel framework integrating hierarchical fusion and attention mechanisms to prioritize semantic and emotional features in therapy transcripts. Our approach combines multiple emotion lexicons, including NRC Emotion Lexicon, VADER, WordNet, and SentiWordNet, with state-of-the-art LLMs such as Flan-T5, LLAMA 2, DeepSeek-R1, and ChatGPT 4. Therapy session transcripts, comprising over 2,000 samples are segmented into hierarchical levels (word, sentence, and session) using neural networks, while hierarchical fusion combines these features with pooling techniques to refine emotional representations. Attention mechanisms, including multi-head self-attention and cross-attention, further prioritize emotional and contextual features, enabling temporal modeling of emotional shifts across sessions. The processed embeddings, computed using BERT, GPT-3, and RoBERTa are stored in the Facebook AI similarity search vector database, which enables efficient similarity search and clustering across dense vector spaces. Upon user queries, relevant segments are retrieved and provided as context to LLMs, enhancing their ability to generate empathetic and contextually relevant responses. The proposed framework is evaluated across multiple practical use cases to demonstrate real-world applicability, including AI-driven therapy chatbots. The system can be integrated into existing mental health platforms to generate personalized responses based on retrieved therapy session data. Experimental results show that our framework enhances empathy, coherence, informativeness, and fluency, surpassing baseline models while improving LLMs' emotional intelligence and contextual adaptability for psychotherapy.

## Introduction

Mental health disorders represent a significant global challenge, impacting approximately 450 million individuals worldwide and resulting in an estimated $1 trillion in productivity losses annually (D. Arias, Saxena, & Verguet, 2022; Brown et al., 2020). In the United States alone, nearly 51.5 million adults experience mental illness each year, underscoring the critical need for accessible and effective mental health care solutions (Adams & Nguyen, 2022). Central to these mental health issues are psychological emotions, which play a crucial role in shaping behaviors, thoughts, and overall well-being (Gross & Jazaieri, 2014). Conditions such as depression, anxiety, and bipolar disorder profoundly affect emotional states, necessitating timely psychological interventions. However, the availability of professional counseling is often constrained by high costs and limited resources (Coombs, Meriwether, Caringi, & Newcomer, 2021; Haugen, McCrillis, Smid, & Nijdam, 2017).

In recent years, advancements in Artificial Intelligence (AI) and machine learning have revolutionized emotion recognition, enabling more sophisticated analysis of human emotions through modalities such as text, speech, and facial expressions (Dalvi, Rathod, Patil, Gite, & Kotecha, 2021). These technologies have been effectively utilized across various domains, including customer service and healthcare. However, despite significant progress, existing AI models often fall short of generating nuanced, empathetic responses that fully resonate with users, particularly in the context of mental health (Khare, Blanes-Vidal, Nadimi, & Acharya, 2023). For instance, while BERT has shown promise in emotion detection within therapy transcripts, it lacks the capability to produce empathetic and contextually appropriate responses (Kumar & Jain, 2022). Similarly, LLMs like Alpaca and Generative Pre-trained Transformer-4 (GPT-4) have demonstrated potential in mental health prediction tasks but still struggle to capture the full depth of human emotions in response generation (Xu et al., 2024).

Recent research has begun to explore the integration of emotion lexicons with neural networks to enhance the emotional understanding and response generation capabilities of AI models. For example, Nandwani and Verma combined

Convolutional Neural Networks (CNNs) with the NRC (National Research Council Canada) Emotion Lexicon to improve emotion detection from text, achieving deeper emotional insights (Nandwani & Verma, 2021). Likewise, Arias and colleagues employed Recurrent Neural Networks (RNNs) alongside the VADER (Valence Aware Dictionary for Sentiment Reasoning) sentiment analysis tool to analyze social media posts for signs of depression, yielding significant improvements in sensitivity and specificity (F. Arias, Nunez, Guerra-Adames, Tejedor-Flores, & Vargas-Lombardo, 2022). These studies highlight the potential of combining lexicon-based approaches with neural networks. However, existing models still struggle to fully capture the intricacies of human emotions, often leading to responses that do not adequately reflect the user's emotional state.

The advent of LLMs such as GPT-3 (Brown et al., 2020), GPT-4 (Kalyan, 2024), and the Fine-Tuned Language Model (Flan-T5) (Chung et al., 2024) has opened new avenues for understanding and generating human-like text, particularly in mental health applications (Xu et al., 2024). For example, Xu et al. (2023) demonstrated the use of GPT-4 in enhancing mental health question-answering tasks (Rodrigues et al., 2024). Gao et al. explored Flan-T5's effectiveness in predicting mental health outcomes from social media data (Nowacki, Sitek, & Rybiński, 2024). When fine-tuned for specific tasks, these models have shown promise in improving mental health prediction and response generation. However, despite these advancements, integrating emotional lexicons into LLMs for enhanced empathetic response generation remains underexplored. This gap results in models that excel in either detection or generation but fail to synergistically combine these capabilities to fully resonate with users emotionally (Gu et al., 2024; Ma, Mei, & Su, 2023).

Our approach introduces hierarchical fusion strategies by segmenting therapy transcripts into word, sentence, and session-level representations, thereby improving contextual awareness across multi-turn dialogues. Additionally, advanced attention mechanisms, including multi-head self-attention and cross-attention, are employed to emphasize emotionally salient features and capture temporal emotional shifts effectively. To further enrich contextual embeddings, representations are transformed using BERT, GPT-3, and RoBERTa and subsequently stored in a FAISS vector database enabling efficient retrieval of relevant information during dialogue generation. By integrating these techniques, our framework significantly enhances empathy, coherence, informativeness, and fluency in LLM-generated responses. Experimental evaluations demonstrate that Emotion-Aware Embedding Fusion outperforms baseline models, underscoring the effectiveness of hierarchical fusion and attention-enhanced embeddings in advancing the emotional intelligence of LLMs. This work represents a crucial step toward improving AI-driven mental health support and addressing the growing global mental health crisis. The main contributions of this study are as follows:

We propose a hierarchical fusion strategy that segments therapy session transcripts into multiple levels (word, sentence, session) to improve emotional and contextual understanding in LLMs.

We introduce attention-enhanced embedding refinement, integrating multi-head self-attention and cross-attention to prioritize emotionally salient features and model temporal shifts in therapy dialogues.

We enhance contextual retrieval using emotion lexicons and FAISS-based vector search, enabling LLMs to generate empathetic, coherent, and contextually appropriate responses, outperforming baseline models.

The paper outlines the methodology in Section II, experimental results in Section III, and conclusions with future directions in Section IV.

## Proposed Methodology

Our approach evaluates the effectiveness of using lexicon dictionaries with various LLMs to enhance emotional state detection and response in a psychiatric context, as presented in Figure 1.

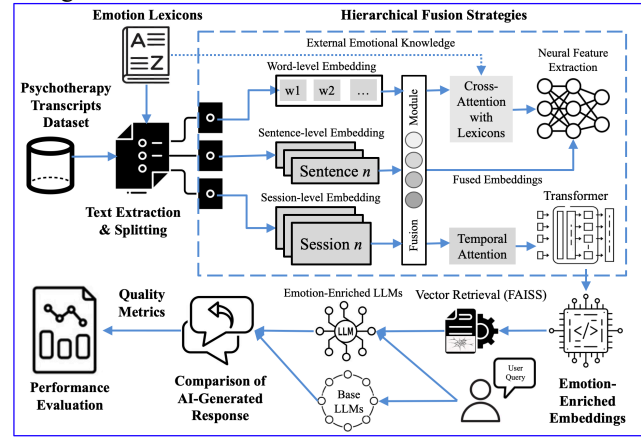

Figure 1: Proposed framework presenting the methodology steps involved in emotion-aware response generation with large language models.

## Dataset

The primary dataset used in this study consists of psychotherapy transcripts from the Counseling and Psychotherapy Transcripts, Client Narratives, and Reference Works database (Shapira et al., 2021). This resource includes over 2,000 transcripts of therapy sessions, patient narratives, and reference works. The dialogues in these transcripts cover a wide range of topics and discussions, including therapeutic interventions, patient histories, emotional disclosures, coping mechanisms, and interactions between therapists and clients. Specific topics discussed include anxiety, depression, relationship issues, trauma, addiction, grief, self-esteem, and personal growth.

**Text Extraction & Splitting**

The extraction process involves removing irrelevant information such as session metadata, timestamps, and non-verbal cues. This allows the focus to be on essential text elements that convey emotional states. For example, irrelevant information is removed by using regex patterns to filter out non-essential text. Segmenting the text into smaller chunks allows for detailed and precise analysis. For instance, the text is split into sentences and further into phrases that capture nuanced emotions.

Let $T$ be the complete transcript, $S_i$ be sentences in $T$, and $P_{ij}$ be phrases within each sentence.

$$T = \{S_1, S_2, \ldots, S_n\}$$
$$S_i = \{P_{i1}, P_{i2}, \ldots, P_{im}\}$$

This segmentation helps in creating manageable chunks, preserving context and enhancing embedding effectiveness. For example, the sentence "I feel anxious about my job" can be split into phrases like "I feel anxious" and "about my job," both retaining the emotional context which is crucial for generating meaningful and empathetic responses. We incorporated the following emotion lexicons to enrich this dataset with emotional cues.

- The NRC Emotion Lexicon is a list of English words associated with eight basic emotions: anger, fear, anticipation, trust, surprise, sadness, joy, and disgust. This lexicon helps identify and categorize emotional expressions within the text, adding a layer of emotional understanding to the analysis (Al Maruf et al., 2024).
- The VADER lexicon is designed for sentiment analysis in social media contexts. VADER assigns a sentiment score to each word, and is particularly effective in analyzing short, informal texts, making it a valuable tool for understanding the sentiment conveyed in conversational language (Hutto & Gilbert, 2014).
- WordNet is a lexical database that groups English words into synsets, representing specific concepts and their semantic relations. It aids in understanding word meanings and context in text analysis (Miller, 1995).
- SentiWordNet extends WordNet by assigning sentiment scores (positive, negative, or neutral) to synonym sets, making it useful for sentiment analysis in various contexts (Esuli & Sebastiani, 2006).

**Embedding Transformation**

For embedding transformation, we use default embeddings like BERT (Qin et al., 2023) and RoBERTa (Luo, Phan, & Reiss) from Hugging Face, and embeddings from OpenAI are GPT-3 and Codex (Hadi et al., 2023). These embeddings convert each transcript segment into high-dimensional vectors, capturing essential semantic and emotional nuances crucial for effective vector retrieval and response generation.

The embeddings serve as numerical representations of transcript segments, allowing models to deeply process and understand the text. By converting text into high-dimensional vectors, models capture and analyze semantic and emotional content, enabling nuanced text analysis and accurate response generation.

Given a transcript segment $x$, it is transformed into an embedding $\phi(x)$ using an embedding model:

$$\phi(x) = \text{EmbeddingModel}(x) \quad (1)$$

where $EmbeddingModel$ could be any specific LLM. The transformation $\phi(x)$ maps the transcript segment $x$ into a high-dimensional vector space.

Assuming $x$ is represented by a sequence of tokens $(x_1, x_2, \ldots, x_n)$, each token $x_i$ is embedded into a vector $\mathbf{e}_i$. To capture contextual dependencies, a self-attention mechanism computes a weighted sum of the embeddings:

$$\mathbf{a}_i = \sum_{j=1}^{n} \alpha_{ij} \mathbf{e}_j \quad (2)$$

where $\alpha_{ij}$ are the attention weights calculated as (Huang, Liang, Qin, Zhong, & Lin, 2023):

$$\alpha_{ij} = \frac{\exp(\mathbf{e}_i \cdot \mathbf{e}_j)}{\sum_{k=1}^{n} \exp(\mathbf{e}_i \cdot \mathbf{e}_k)} \quad (3)$$

This results in a context-aware embedding for each token, aggregated to form the final representation:

$$\phi(x) = \sum_{i=1}^{n} \mathbf{a}_i \quad (4)$$

**Vector Retrieval**

We store the embedded vectors in an efficient vector database designed for high-dimensional data, specifically using FAISS (Facebook AI Similarity Search). FAISS enables fast similarity search and clustering of dense vectors, which is essential for our large-scale machine learning tasks(Ghadekar, Mohite, More, Patil, & Mangrule, 2023). In our study, FAISS stores and retrieves embeddings efficiently.

Given an input query $q$, FAISS retrieves the most relevant transcript segments $\{x_i\}$ by computing the cosine similarity between the query embedding $\phi(q)$ and the stored embeddings,

$$\{\phi(x_i)\}: \text{sim}(q, x_i) = \frac{\phi(q) \cdot \phi(x_i)}{\|\phi(q)\| \|\phi(x_i)\|} \quad (5)$$

For example, if a user queries about feeling anxious, FAISS finds the closest matching transcript segments related to anxiety. This helps the model generate a response that is relevant and empathetic to the user's concern.

**Response Generation**

The selected LLMs generate responses based on retrieved segments. Vector retrieval provides relevant transcript segments $\{x_i\}$ for a given input query $q$. The response generation process is formulated as:

$$r = \text{LLM}(q, \{x_i\}) \tag{6}$$

where $r$ is the generated response, $q$ is the input query, and $\{x_i\}$ are the retrieved segments.

Lexicon resources enhance the models' empathetic and coherent response capabilities by providing emotional cues and semantic meanings. For example, the system retrieves relevant segments and generates a supportive response if a user queries about feeling anxious. Algorithm 1 shows the response generation of the proposed framework. It starts by initializing $t$ to 0 and iterates through each transcript segment $x_t$ in $X$. For each segment, it computes the embedding $\psi(x_t)$ using the embedding model $\mathcal{E}$. If elements from the lexicon $\mathcal{L}$ are present in $x_t$, the algorithm enhances the embedding by computing $\psi_\mathcal{L}$ with $\delta(e_i, \mathcal{L})$ and updates $\psi(x_t)$. If no such elements are found, $\psi(x_t)$ remains unchanged. After processing all segments, the algorithm increments $t$ and repeats the process. Finally, it applies the function $\Psi$ to the set of segments $X$ and their embeddings $\psi(X)$, generating the emotion-aware responses $\mathcal{R}$. The function $\Psi$ processes the transcript segments and their embeddings to produce the final set of emotion-aware responses stored in $\mathcal{R}$. The following LLMs are used to generate responses.

---

Algorithm 1: Response generation mechanism.

**Input:** Transcript segments X, lexicon dictionaries $\mathcal{L}$, enhancement function $\delta(e_i, \mathcal{L})$, embedding of segment $\psi(x_t)$, enhanced embedding with lexicon features $\psi_\mathcal{L}$
**Parameter**: Embedding model $\mathcal{E}$, similarity threshold $\tau$
**Output:** Emotion-aware responses.
1: Initialize t←0.
2: **while** X≠∅ do
3:     Extract segment $x_t$ from X.
4:     Compute embedding $\psi(x_t) \leftarrow \mathcal{E}(x_t)$.
5:     **if** $\exists e \in \mathcal{L}$ where $e \in x_t$ then
6:         Enhance $x_t$ with $\psi_\mathcal{L} \leftarrow \sum_{i=1}^{|x_t|} \delta(e_i, \mathcal{L})$.
7:         Update embedding $\psi(x_t) \leftarrow \psi(x_t) + \psi_\mathcal{L}$.
8:     **else**
9:         Continue with $\psi(x_t)$.
10:    **end if**
11:    $t \leftarrow t+1$
12: **end while**
13: **return** $\mathcal{R} \leftarrow \Psi(X, \psi(X))$

---

**Flan-T5 Large**
A well-known model for natural language understanding tasks, optimized for text-to-text transformations. In this model, the probability of generating an output sequence $y$ given an input sequence $x$ is represented as (Chung et al., 2024):

$$P(y \mid x) = \prod_{t=1}^{T} P(y_t \mid y_{<t}, x; \theta) \tag{7}$$

where $y$ is the output sequence, $x$ is the input sequence, and $\theta$ represents the model parameters.

The derivation involves modeling the conditional probability of each token $y_t$ in the output sequence, given the previous tokens $y_{<t}$ and the input sequence $x$.

**Llama 2 13B**
It's a large-scale model with robust understanding and generation capabilities, particularly effective in handling longer contexts and providing nuanced responses. Its autoregressive generation process is represented as (Touvron et al., 2023):

$$P(x_t \mid x_{<t}) = \text{softmax}(W_h h_t) \tag{8}$$

where $x_t$ is the current token, $x_{<t}$ are the preceding tokens, $W_h$ is a weight matrix, and $h_t$ represents the hidden state at time $t$.

This captures the dependencies and context within the input sequence, allowing the model to generate coherent and contextually appropriate text.

**ChatGPT 3.5**
It is a widely used model with general conversational abilities, expected to show marked improvements in empathy and coherence with lexicons due to its conversational design. The decoding process is modeled by (Brown et al., 2020):

$$P(y_t \mid y_{<t}, x) = \text{softmax}(W_y y_{t-1} + W_x x + b) \tag{9}$$

where $y_t$ is the generated token at time $t$, $y_{<t}$ are the preceding tokens in the output sequence, $x$ is the input sequence, $W_y$ and $W_x$ are weight matrices, and $b$ is a bias term.

This formulation allows the model to generate each token $y_t$ based on the preceding context and the input sequence.

**ChatGPT 4**
The latest iteration in the GPT series, has advanced training on diverse datasets and is expected to provide balanced improvements across all metrics with different lexicons. Its autoregressive nature of generation is modeled as (Liu et al., 2023):

$$P(x_t \mid x_{<t}) = \text{softmax}(W_t h_{t-1} + b_t) \tag{10}$$

where $x_t$ is the token at position $t$, $x_{<t}$ are the preceding tokens, $W_t$ is a weight matrix specific to the position $t$, $h_{t-1}$ is the hidden state from the previous step, and $b_t$ is a bias term.

This equation models the probability distribution of the current token based on the hidden states derived from preceding tokens, ensuring the generation of contextually coherent responses.

## Quality Metrics

We define several metrics to evaluate the quality of the chatbot's responses. Each metric assigns a score based on specific criteria. The functions are:

### Empathy Score

It evaluates how empathetic a response is. Instead of a simple modulo operation, we use a more complex function to capture the nuances of empathy in language. The score is calculated as follows (Lima & Osório, 2021):

$$E = \frac{\sum_{i=1}^{N} \text{weight}_i \cdot \text{emotion}_i}{\sum_{i=1}^{N} \text{weight}_i} \tag{11}$$

where $\text{weight}_i$ represents the importance of the $i$-th emotional word, and $\text{emotion}_i$ is the intensity of the emotion conveyed by the $i$-th word in the response. This weighted sum captures the overall empathetic content more effectively.

### Coherence Score

It assesses the coherence of a response by measuring the logical flow and consistency of the text. It is determined by (Marchenko, Radyvonenko, Ignatova, Titarchuk, & Zhelezniakov, 2020):

$$C = \sum_{i=1}^{N-1} \exp\left(-\frac{d(w_i, w_{i+1})}{\sigma^2}\right) \tag{12}$$

where $d(w_i, w_{i+1})$ is the semantic distance between consecutive words $w_i$ and $w_{i+1}$, and $\sigma$ is a scaling factor that adjusts the sensitivity to semantic distances. This formulation uses the exponential function to penalize significant semantic gaps, ensuring a coherent response.

### Informativeness Score

It measures how informative a response is by considering the amount and relevance of information provided. The score is given by (Senbel, 2021):

$$I = \log\left(1 + \sum_{i=1}^{N} \text{tf} - \text{idf}(w_i)\right) \tag{13}$$

where $tf - idf(w_i)$ is the term frequency-inverse document frequency of word $w_i$. This logarithmic function accounts for the diminishing returns of adding more information, emphasizing the importance of key terms.

### Fluency Score

It evaluates the fluency of a response, ensuring it reads naturally. The score is calculated as (Villalobos, Torres-Simón, Pacios, Paul, & Del Río, 2023):

$$F = \frac{1}{N} \sum_{i=1}^{N} P(w_i \mid w_{i-1}, w_{i-2}, \ldots, w_{i-n}) \tag{14}$$

where $P(w_i \mid w_{i-1}, w_{i-2}, \ldots, w_{i-n})$ is the conditional probability of word $w_i$ given its $n$-gram history.

This average probability captures the smoothness and naturalness of the language used. To calculate the average score for a given metric, the following formula is used

$$A = \frac{1}{M} \sum_{i=1}^{M} metric\_function(\text{response}_i) \tag{15}$$

where $M$ is the total number of responses, and $metric\_function$ is one of the defined metric functions. The '$metric\_function$' dictionary maps the four critical evaluation criteria to their respective scoring functions.

The overall performance score for each model is calculated as:

$$\text{Score}_{\text{avg}} = \frac{1}{4}(E + C + I + F) \tag{16}$$

This provides insights into improvements achieved by incorporating NRC and VADER datasets into LLMs for psychiatric applications. Each metric is scored on a scale from 1 to 5, with higher scores indicating better performance. For evaluation, we offered two approaches for sophisticated comparisons of our proposed study. The evaluation is conducted ($i$) with and ($ii$) without the emotion lexicon resources to determine the impact of emotional cues on the models' performance.

## Results Evaluations

### Baseline Performances

In our initial evaluation, we assessed the baseline performance (without lexicon adding) of four state-of-the-art LLMs in the context of psychotherapy-related tasks. Each model was evaluated across four key metrics: empathy, coherence, informativeness, and fluency. The results are summarized in Table 1.

| Model | Empathy | Coherence | Informativeness | Fluency |
|---|---|---|---|---|
| Flan-T5 Large | 3.5 | 2.0 | 3.0 | 4.0 |
| Llama 2 13B | 2.0 | 3.0 | 4.0 | 5.0 |
| ChatGPT 3.5 | 4.0 | 5.0 | 1.0 | 2.0 |
| ChatGPT 4 | 5.0 | 2.0 | 2.0 | 3.0 |

Table 1: Multi-metric comparisons of baseline LLMs.

ChatGPT 4 led in empathy (5.0), benefiting from advanced training on diverse datasets, while ChatGPT 3.5 also performed well (4.0), reflecting its design for emotional engagement. However, Llama 2 13B (2.0) and Flan-T5 Large (3.5) showed weaker empathy, likely due to their architecture's focus on other aspects like context handling or general text tasks. ChatGPT 3.5 excelled (5.0) in coherence, maintaining logical consistency in dialogues, whereas ChatGPT 4 and Flan-T5 Large (2.0) underperformed, possibly due to architectural trade-offs. Llama 2 13B was the most informative (4.0) due to its detailed response generation capabilities. Flan-T5 Large and ChatGPT 4 (3.0 and 2.0) provided less depth, focusing more on fluency or empathy. ChatGPT 3.5's low informativeness (1.0) indicates its priority on emotional resonance over detailed content.

**Affect-Enriched LLMs Comparisons**

We tested the same four LLMs on psychotherapy-related metrics to evaluate the impact of incorporating NRC lexicons to enrich the effects of embedding. The results based on NRC affect enrichment are summarized in Table 2.

| Model | Empathy | Coherence | Informativeness | Fluency |
|---|---|---|---|---|
| Flan-T5 Large | 5.0 | 1.0 | 2.0 | 3.0 |
| Llama 2 13B | 1.0 | 2.0 | 3.0 | 4.0 |
| ChatGPT 3.5 | 5.0 | 1.0 | 2.0 | 3.0 |
| ChatGPT 4 | 5.0 | 1.0 | 2.0 | 3.0 |

Table 2: Performance of LLMs after embedding enrichment with NRC lexicon.

Incorporating NRC lexicons led to significant changes in performance across all models. Flan-T5 and ChatGPT 3.5 both saw significant increases in empathy, improving from 3.5 to 5.0 (an increase of approximately 43%) and 4.0 to 5.0 (25%), respectively. However, this boost in empathy came with a substantial decrease in coherence, dropping to 1.0 in both cases (a decrease of 50% for Flan-T5 and 80% for ChatGPT 3.5), indicating that while these models became more emotionally responsive, they struggled to maintain logical consistency.

ChatGPT 4 also maintained its top empathy score at 5.0, similar to ChatGPT 3.5. However, like the other models, its coherence dropped to 1.0, representing a 50% decrease from its baseline of 2.0. The nearly identical performance of ChatGPT 3.5 and ChatGPT 4 can be attributed to their shared architectural foundation and similar design priorities, which emphasize empathy at the expense of other metrics. Both models likely leverage the same core mechanisms for emotion recognition and response generation, leading to parallel outcomes when enhanced with NRC lexicons. Llama 2, which initially struggled with empathy (2.0), experienced a sharp decline to 1.0 (a 50% decrease), suggesting that the NRC integration did not enhance its ability to process emotional cues effectively. However, it retained a better balance in fluency (4.0) and informativeness (3.0), though both metrics saw slight reductions of 20% and 25%, respectively, compared to its baseline.

While NRC integration significantly enhanced empathy across most models, this improvement often came at the cost of coherence. The comparison with general performance highlights a trade-off where models become more emotionally attuned but less capable of maintaining logical, informative, and fluent conversations.

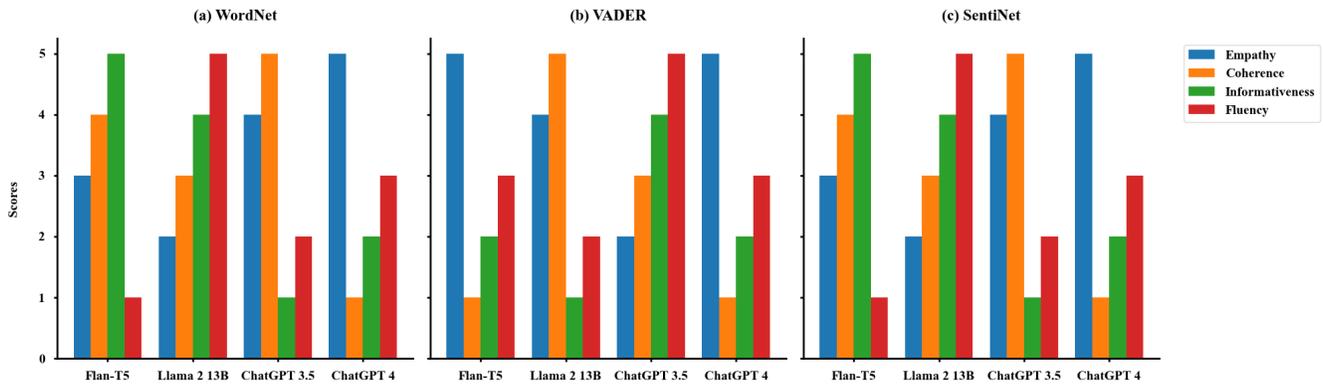

Figure 2: Language models' performance comparison after affect-enriched embeddings with (a) WordNet, (b) VADER, and (c) SentiNet lexicons.

| Models with lexicons | Empathy | Coherence | Informativeness | Fluency |
|---|---|---|---|---|
| Flan-T5 (VADER) vs. WordNet | +67% | -75% | -60% | +200% |
| Flan-T5 (VADER) vs. SentiNet | +67% | -75% | -60% | +200% |
| ChatGPT 3.5 (VADER) vs. WordNet | +50% | -50% | +400% | +60% |
| ChatGPT 3.5 (VADER) vs. SentiNet | +50% | -50% | +400% | +60% |
| Llama 2 (VADER) vs. WordNet | +50% | -40% | -50% | -60% |
| Llama 2 (VADER) vs. SentiNet | +50% | -40% | -50% | -60% |

Table 3: Difference in performance of Multi-metrics between VADER and other lexicons (WordNet, SentiNet) for various LLMs.

Figure 3: Comparisons of generated responses from different LLMs with and without affect-enriched embeddings using NRC lexicon.

The performance of LLMs is notably influenced by the choice of lexicon, as seen in Figure 2 and Table 3. Each lexicon impacts the key metrics differently across the models. To avoid redundancy and overlap, we have presented a few random comparisons in Table 3, in which plus values indicate improvement and minus entries present decreased performance.

Flan-T5's empathy improves by 67% with VADER due to its refined sentiment analysis. However, this increase in emotional sensitivity causes a 75% decline in coherence, disrupting the model's logical flow. ChatGPT 3.5 shows a 400% boost in informativeness with VADER, but this enhancement leads to a 50% drop in coherence and fluency, highlighting the model's struggle to balance detailed content with conversational quality. Llama 2 maintains high fluency with WordNet and SentiNet but drops by 60% with VADER, suggesting that VADER's emphasis on empathy can interfere with producing smooth dialogue. Across all models, ChatGPT 4 consistently achieves high empathy scores (5.0) across all lexicons, demonstrating its robustness in emotional understanding. However, the consistent drop in coherence to 1.0 indicates that, regardless of the lexicon, there is a significant trade-off between high empathy and maintaining logical consistency in the conversation. This suggests that ChatGPT 4's architecture, while strong in detecting and responding to emotions, struggles to balance this with producing coherent narratives.

**LLMs Responses Comparison**

We have demonstrated the comparative analysis of generated responses from four LLMs with and without affect-enriched embeddings. Figure 3 presents this comparison based on the enrichment of one lexicon (NRC Emotion Lexicon) with the same question. The questionnaire-based performance reveals the significant impact of integrating the lexicon on the response generation by LLMs.

- Llama 2 13B shows a significant improvement with the NRC dataset. Without it, the model provides a standard, albeit somewhat shallow, response that acknowledges the user's feelings but lacks detailed guidance. With the NRC dataset, the model's response becomes more empathetic and actionable, offering practical advice like talking to a supervisor or documenting the incident. This enhancement highlights Llama 2 13B's increased ability to process emotional content effectively, making the interaction more supportive and useful.
- Flan-T5 struggles to generate emotionally nuanced responses without the NRC dataset, often offering basic acknowledgments like "He praised his colleague," which lack depth. While the NRC dataset helps the model adopt a more empathetic tone, the responses remain fragmented and fail to engage with the user's emotional nuances fully. This issue is partly due to Flan-T5's tokenization mechanism, which is capped at 512 tokens. When inputs exceed this limit, the model may truncate the text, leading to incomplete or disjointed responses. Additionally, the special tokens used to capture specific emotions can disrupt the flow of the response, further limiting its effectiveness.
- ChatGPT 3.5 delivers a coherent and supportive response even without the NRC dataset, though it remains somewhat generic. The model recognizes the user's frustration and suggests speaking with a colleague or supervisor. When the NRC dataset is applied, the response becomes more emotionally attuned, reflecting a better understanding of the user's anger, though the advice itself does not significantly change. This indicates that the NRC dataset enhances the tone of the response more than the content.
- ChatGPT 4 also benefits from the NRC dataset. Initially, without it, the responses are mechanical, simply recognizing that the client is upset. With the NRC dataset, the model adopts a more empathetic tone, providing thoughtful advice and encouraging positive actions. This shift suggests that ChatGPT 4, like ChatGPT 3.5, enhances its emotional engagement with the integration of the NRC dataset, improving the overall supportiveness of the interaction.

The comparative analysis underscores the value of integrating affect-enriched embeddings into LLMs, particularly for applications requiring emotional sensitivity and coherence. While the NRC dataset significantly improves the empathetic tone of responses, the results also highlight the challenges of balancing emotional engagement with the coherence and informativeness of the output.

## Discussion

The integration of emotion lexicons with LLMs has demonstrated significant potential in enhancing the empathy and contextual understanding of AI-driven responses within psychotherapy applications. Our study reveals that while LLMs like Flan-T5, ChatGPT 3.5, and ChatGPT 4 excel in emotional engagement, the incorporation of lexicons such as NRC, VADER, WordNet, and SentiNet introduces a complex trade-off between empathy and coherence.

A primary challenge was the token limitation in Flan-T5 Large and Llama 2 13B, leading to indexing errors and difficulty processing longer inputs, which are crucial in psychotherapy. Addressing this through model truncation or attention optimization could improve the handling of extended dialogues. Another challenge was balancing informativeness with other metrics like empathy and coherence. For instance, ChatGPT 3.5's 400% improvement in informativeness with the VADER lexicon led to a 50% drop in coherence and fluency (Table 3). This highlights the trade-off between providing detailed information and maintaining conversational quality, which is essential in therapeutic settings.

Balancing empathy with coherence and informativeness also proved difficult. Integrating NRC lexicons improved empathy scores (Table 2), but often reduced coherence, as seen in models like ChatGPT 4. This suggests that while LLMs can be tuned for emotional engagement, maintaining balance across metrics remains a challenge.

Additionally, some models showed good metric scores but failed to generate coherent responses, indicating a gap between quantitative metrics and actual performance. Llama 2, for example, generated multiple outputs for a single response on the NRC dataset, emphasizing the need for better context handling and response generation. Informativeness was particularly challenging, as models often sacrificed coherence and fluency to provide more detail. This underscores the need for advanced metrics that capture the nuances of empathy and coherence without compromising informativeness.

These findings underline the need for a more nuanced approach to integrating emotional embeddings in AI models for mental health applications. While emotional sensitivity is crucial, maintaining a balance of coherence and informativeness is essential to generate meaningful and supportive responses. This balance is particularly critical in therapeutic settings where logical consistency and clarity are as important as emotional resonance.

## Conclusion

This study advances AI-driven psychotherapy by integrating emotion lexicons with state-of-the-art LLMs. While the results highlight significant improvements in empathy and contextual understanding, they also reveal critical challenges, particularly in balancing empathy with coherence and informativeness. The trade-offs observed in models such as Flan-T5 and ChatGPT 3.5 and the token limitations in Llama 2 13B underscore the complexity of developing emotionally intelligent AI systems. To address these challenges, model fine-tuning specifically for psychiatric use cases, with a balanced dataset, is recommended. Additionally, implementing long-context handling techniques and refining evaluation metrics to capture these models' nuanced performance better are essential steps forward. Future work should also focus on expanding the dataset to include more diverse emotional contexts and conducting human evaluations to gain deeper insights into the models' performance in real-world applications.